%% file: paper_arxiv.tex
\ifdefined\TEAMAnonymousReview
  \documentclass[sigconf,anonymous,review]{acmart}
\else
  \documentclass[sigconf,nonacm]{acmart}
\fi

\AtBeginDocument{}
\usepackage{mathtools,amsthm}
\usepackage{multirow}
\usepackage{array}
\usepackage{placeins}
\usepackage[normalem]{ulem}
\input{macros.tex}

\setcopyright{none}
\acmDOI{}
\acmISBN{}

\begin{document}
\title{TEAM: Topology-Aware, Semantic-Sensitive, General Transformer for Multimodal-Attributed Graphs}
\input{authors_arxiv.tex}
\input{sections/abstract.tex}
\maketitle
\pagestyle{empty}
\input{sections/introduction.tex}
\input{sections/preliminaries.tex}
\input{sections/methods.tex}
\input{sections/experiments.tex}
\clearpage
\input{sections/ethical_considerations.tex}
\vspace{-0.75\baselineskip}
\bibliographystyle{ACM-Reference-Format}
\bibliography{references}
\end{document}

%% file: macros.tex
\theoremstyle{plain}
\newtheorem{theorem}{Theorem}

\newtheorem{lemma}{Lemma}

\theoremstyle{definition}

\theoremstyle{remark}

%% file: authors_arxiv.tex
\makeatletter
\newcommand{\TEAMendauthorgroup}{%
  \global\@insideauthorgroupfalse
  \g@addto@macro\addresses{\let\and\@typeset@author@bx}}
\makeatother
\author{Yushuo Li\texorpdfstring{\textsuperscript{1}}{}}
\TEAMendauthorgroup
\author{Xunkai Li\texorpdfstring{\textsuperscript{1}}{}}
\TEAMendauthorgroup
\author{Zhengyu Wu\texorpdfstring{\textsuperscript{1}}{}}
\TEAMendauthorgroup
\author{Ronghua Li\texorpdfstring{\textsuperscript{1*}}{}}
\TEAMendauthorgroup
\author{Guoren Wang\texorpdfstring{\textsuperscript{1}}{}}
\TEAMendauthorgroup
\thanks{\textsuperscript{1}Beijing Institute of Technology, China.}
\thanks{\textsuperscript{*}Corresponding author: Ronghua Li
  (\href{mailto:lironghuabit@126.com}{lironghuabit@126.com}).}
\renewcommand{\shortauthors}{Li et al.}

%% file: sections/abstract.tex
\begin{abstract}
Multimodal-Attributed Graphs (MAGs) organize heterogeneous data in graphs whose
nodes are characterized by modalities such as text and images rather than
traditional unimodal attributes.  However, many representative MAG methods
organize propagation around node-level states and topology-defined
neighborhoods, which can obscure fine-grained modality contributions: a bounded
neighborhood may contain semantically weak evidence, while compatible context
need not be confined to it.  To address
both limitations, we propose \textbf{TEAM}, a modality-centric framework that
constructs and encodes graph context with node--modality tokens rather than
treating each node as a single interaction unit.
Concretely, TEAM unfolds the local topology into a Modality Encoding Tree and
performs tree-anchored semantic evidence acquisition, organizing local
structural context while expanding the accessible evidence space outside the
constructed local Tree.  We further introduce the Modality-Source Context
Transformer (MSCT), which encodes the selected node--modality tokens and coordinates
interactions among modality--source groups.  By retaining textual
and visual attributes as separate interaction units until readout, it supports
fine-grained cross-node cross-modal exchange without first collapsing them into
whole-node features.  Across the evaluated baselines, TEAM attains the highest
reported mean on all graph-learning metrics, all three G2Text metrics, and both
G2Image metrics.  On Toys, it surpasses the strongest baseline by 3.37 and 2.93
percentage points in Accuracy and Macro-F1, respectively.
\end{abstract}

%% file: sections/introduction.tex
\section{Introduction}

Graph-structured data is ubiquitous in complex systems.
Graph Neural Networks (GNNs) dominate graph representation learning by encoding
structural inductive biases~\citep{Wu_2021}, but conventional pipelines
represent each node with a single vector and thus obscure modality-specific
semantics.  Real-world entities, however, often contain heterogeneous
modalities (e.g., text, images, and audio), forming \emph{Multimodal-Attributed
Graphs} (MAGs)~\citep{ektefaie2023multimodallearninggraphs,
peng2024learningmultimodalgraphssurvey}.  Joint modeling is essential:
multimodal attributes enrich graph reasoning~\citep{zhu2025mosaicmodalitiescomprehensivebenchmark},
while topology contextualizes multimodal content~\citep{fang2025graphgptosynergisticmultimodalcomprehension,
jin2024instructg2isynthesizingimagesmultimodal}.  Yet, combining unstructured
signals with structural constraints remains challenging because of their
semantic gap.

Despite the notable progress in multimodal graph learning, existing approaches suffer from two key limitations.
\textbf{(L1) Node-centric Granularity.}
Because graph propagation treats nodes as atomic units, most methods follow one of two designs.
Early fusion compresses modalities before propagation, obscuring modality-specific semantics~\citep{he2025unigraph2learningunifiedembedding}.
Conversely, late fusion propagates modality-specific representations over separate graphs and merges them afterward~\citep{wei2019mmgcn,hu2024modalityindependentgraphneuralnetworks}.
Both retain nodes as the smallest propagation unit, making direct
node--modality token interaction across nodes unavailable in these formulations and
limiting the use of complementary signals such as a neighbor's image for a
target's text~\citep{li2025representationlearningmutualinfluence}.
\textbf{(L2) Topology-Semantic Misalignment.}
Standard methods adopt a topology-first perspective, aggregating through rigid message passing (e.g., $k$-hop) assuming structural neighbors are useful~\citep{wei2019mmgcn,tao2020mgat,hu2024modalityindependentgraphneuralnetworks}.
Yet, proximity does not ensure semantic relevance: a bounded neighborhood may contain task-irrelevant modalities, whereas compatible evidence can lie outside it.
These designs leave unresolved how to jointly preserve local structural
organization, select individual modalities, and supplement bounded out-of-tree
evidence.

\begin{figure*}[t!]
\centering
\begin{minipage}[b]{0.485\textwidth}
\centering
\includegraphics[width=\linewidth]{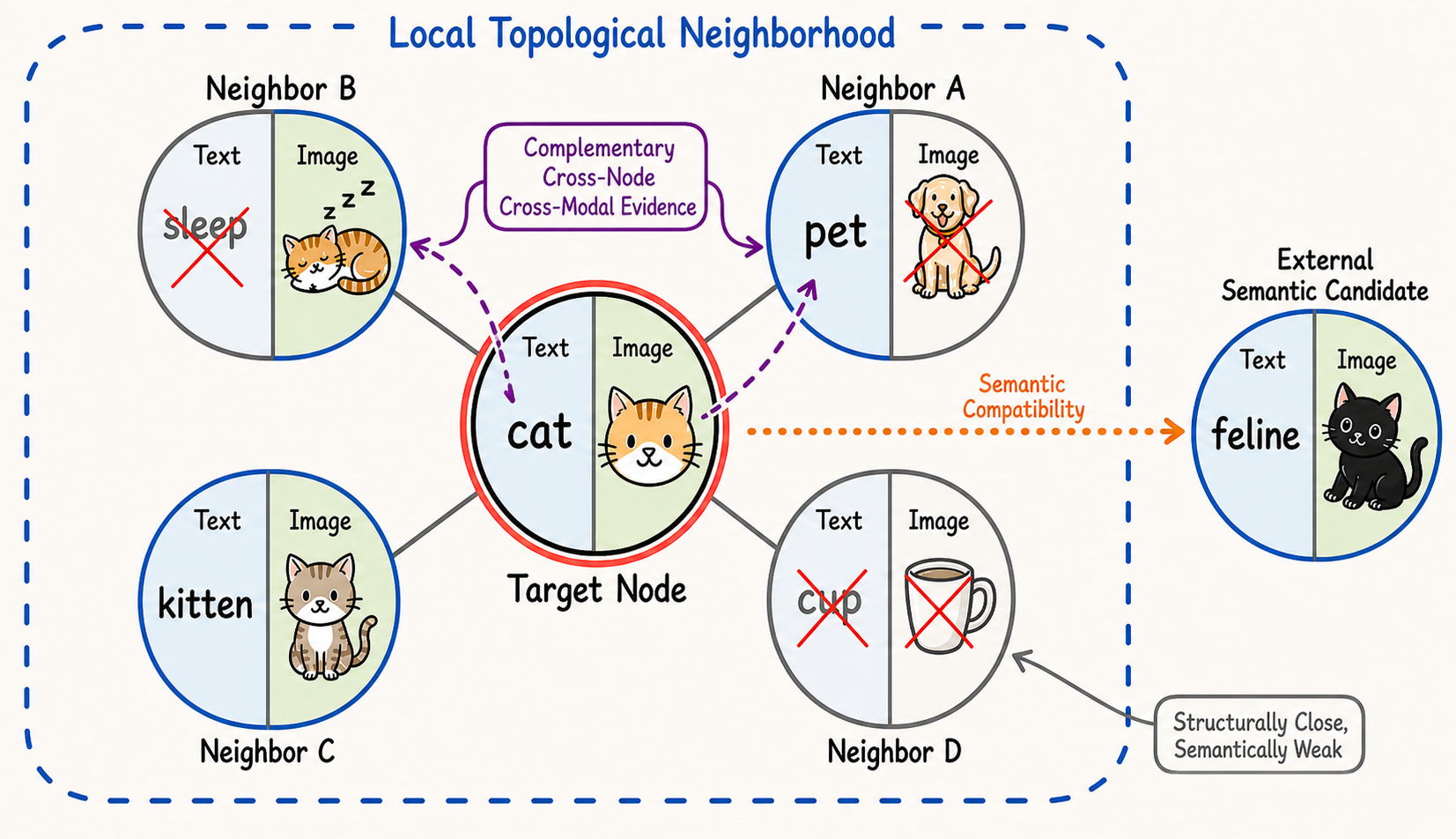}
\caption{\textbf{Structural Proximity $\neq$ Semantic Relevance.} A local
topological neighborhood can contain weak modalities while compatible evidence
may remain outside the topology-defined local context.}
\label{fig:toy_example}
\end{minipage}
\hfill
\begin{minipage}[b]{0.485\textwidth}
\centering
\includegraphics[width=\linewidth]{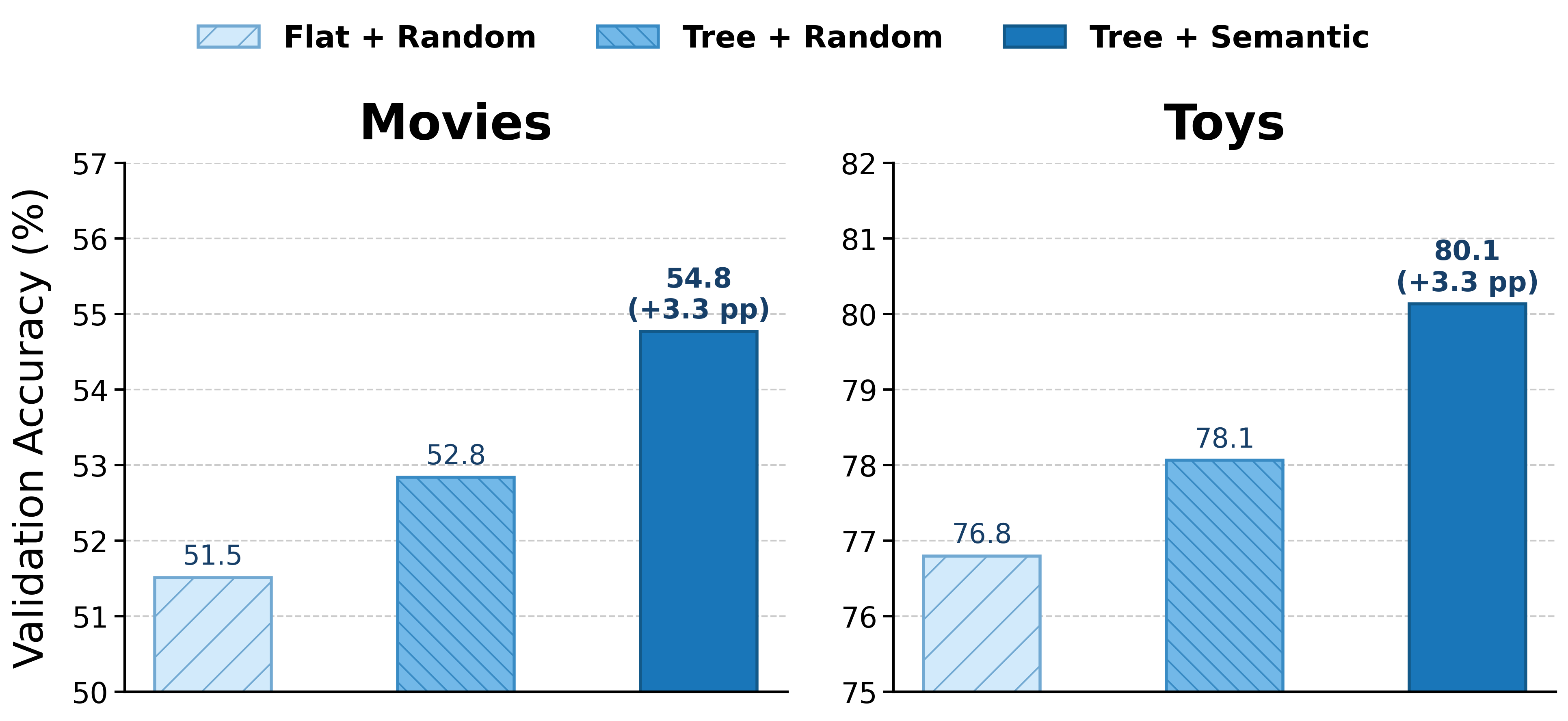}
\caption{\textbf{Controlled Validation Study of Structural Organization and Semantic
Selection.} All variants draw from the same depth-limited neighborhood pool.
Flat+Random and Tree+Random wire identical random contexts as a target-centered
star or along BFS parent--child edges; Tree+Semantic uses the latter wiring and
multimodal relevance ranking (displayed gains: versus Flat+Random).}
\label{fig:empirical_study}
\end{minipage}
\Description{Two motivating panels.  The left panel shows a target node and its
local multimodal neighborhood.  Some nearby modalities are weak, while a
semantically compatible candidate remains outside the topology-defined local context.
The right panel compares flat random context, tree-organized random context,
and tree-organized semantic context on Movies and Toys.}
\end{figure*}

Figures~\ref{fig:toy_example} and~\ref{fig:empirical_study} motivate how graph
context should be represented and selected.  Figure~\ref{fig:toy_example}
shows that useful and irrelevant signals may coexist in different modalities
of a neighboring node, while compatible modality evidence may occur beyond the
local context.  Using the same GCN encoder, the validation results in
Figure~\ref{fig:empirical_study}
isolate two construction choices: BFS parent--child connectivity improves over
flat wiring under identical random contexts, and multimodal relevance ranking
further improves the same tree organization.
Together, these observations motivate modality-level context construction that
preserves local structural organization, prioritizes semantically relevant
evidence, and broadens evidence acquisition beyond topology-defined local
context within one coordinated construction process.

\begingroup
\setlength{\emergencystretch}{1em}
To this end, we propose \textbf{TEAM}
(\textbf{T}opology-aware,\allowbreak{} s\textbf{E}mantic-sensitive,\allowbreak{}
gener\textbf{A}l transfor\textbf{M}er), a modality-centric model for organizing
graph structure, selecting semantic evidence, and encoding cross-modal context.
TEAM organizes the local neighborhood as a Modality Encoding Tree (MET), where
textual and visual attributes are represented as separate tokens with explicit
branch and hop information.
Branch-aware allocation and semantic refill determine query-specific local
node--modality token anchors, while selective semantic retrieval adds compatible
external node--modality tokens under a bounded budget.  Two query-specific streams of
our Modality-Source Context Transformer (MSCT) encode the resulting sequences.
Within each MSCT layer, token self-attention and feed-forward transformation are
followed by modality--source grouping, group interaction, and matched write-back
to the corresponding tokens within each query sequence.  The MSCT output
readout fuses the two aligned query roots, uses the fused representation to query
the final modality--source group memories, and produces the node embedding.
Together, MET organizes topology,
the sampler selects modality-level evidence, and MSCT retains modality and
source through interaction and readout.
\par
\endgroup

\textbf{Our Contributions}: (1) \textbf{\textit{\uline{Problem
Formalization.}}} We formalize two coupled limitations of multimodal graph
learning: node-centric granularity restricts fine-grained cross-node
cross-modal reasoning, while locality-constrained aggregation may retain
irrelevant modality evidence and restrict semantic coverage; (2)
\textbf{\textit{\uline{Novel Modality-centered Framework.}}} We propose TEAM, which
combines a Modality Encoding Tree with tree-anchored modality-level context
sampling and MSCT, preserving
topological guidance while enabling modality-level selection,
Tree-external evidence acquisition, and cross-modal interaction; and
(3) \textbf{\textit{\uline{SOTA Performance.}}} Extensive experiments
demonstrate that TEAM achieves state-of-the-art performance on all evaluated
graph-learning and multimodal tasks, together with the strongest robustness
under the tested corruptions on Grocery and Toys.

%% file: sections/preliminaries.tex
\section{Preliminaries and Related Work}

\subsection{Notation and Problem Formulation}

Let $\mathcal{G}=(\mathcal{V},\mathcal{E},\{\mathbf{X}^{m}\}_{m=1}^{M})$
be a multimodal-attributed graph, where $\mathcal{V}=\{v_1,\dots,v_N\}$,
$\mathcal{E}\subseteq\mathcal{V}\times\mathcal{V}$, and
$\mathbf{X}^{m}\in\mathbb{R}^{N\times d_m}$ are the node set, edge set, and
feature matrix of modality $m$, respectively.  For node $v_i$,
$\mathbf{x}_i^m\in\mathbb{R}^{d_m}$ denotes its feature in modality $m$.

\textbf{Problem Statement.}
Given a MAG $\mathcal{G}$ and a target node $v$, we learn a representation
function $f_{\theta}$ that (1) derives \emph{modality-specific representations}
$\{\mathbf{h}_v^m\}_{m=1}^{M}$ from the structural context of $v$, and
(2) adaptively fuses them into a \emph{unified node embedding}
$\mathbf{h}_v$ for downstream prediction.
TEAM instantiates this formulation with text and visual attributes and retains
their modality identities throughout context construction and encoding.

\subsection{Multimodal-Attributed Graph Learning}

Existing MAG methods can be grouped by fusion strategy.

\textbf{Early Fusion.}
Early-fusion methods unify modalities before structural propagation.
With an MoE, UniGraph2 \citep{he2025unigraph2learningunifiedembedding} aligns
features from modality-specific encoders before GNN propagation, so graph
reasoning operates on unified node representations.
This alignment simplifies propagation, but fixes cross-modal interaction
before the target's structural context has been observed.

\textbf{Late Fusion.}
Late-fusion methods preserve modality-specific streams and combine them only after graph propagation.
Examples include MMGCN \citep{wei2019mmgcn}, contrastive denoising models \citep{10075502}, and modality-independent GNNs with adaptive depths \citep{hu2024modalityindependentgraphneuralnetworks}, thereby retaining modality distinctions until final integration.
Keeping the streams separate preserves modality-specific evidence, but delays
their direct interaction until neighborhood aggregation is complete.

Fusion timing, however, does not change the information unit exchanged across
the graph.  Early fusion removes modal identity before propagation, whereas
late fusion preserves separate streams but still propagates each stream as a
node-level state during graph propagation.

Beyond these dominant designs, recent methods introduce interaction or
structure-aware mechanisms.
MGAT \citep{tao2020mgat} uses cross-modal guidance during gated neighborhood aggregation, HGNN-IMA \citep{li2025representationlearningmutualinfluence} models cross-modal influence during propagation, and DMGC \citep{guo2025disentanglinghomophilyheterophilymultimodal} disentangles homophilic and heterophilic structural views.
Yet, propagation remains organized primarily around node-level states, limiting
fine-grained cross-node modality interaction during representation learning and
inference.

\subsection{Neighborhood Sampling and Selection}

Neighborhood sampling limits receptive fields or forms compact mini-batches.
GraphSAGE samples fixed-size neighborhoods \citep{hamilton2017inductive};
FastGCN and LADIES use layer-wise importance sampling
\citep{chen2018fastgcnfastlearninggraph,zou2019layerdependentimportancesamplingtraining};
and GraphSAINT and Cluster-GCN sample or partition subgraphs
\citep{zeng2020graphsaintgraphsamplingbased,chiang2019clustergcn}.  These methods
primarily reduce computation at node or subgraph granularity for scalable training.

\newpage
Beyond efficiency, PASS learns task-adaptive sampling \citep{yoon2021pass}, and
AGS-GNN balances feature similarity and diversity \citep{das2024agsgnn}.
Pro-GNN and GNNGuard instead repair topology by refining graphs or re\-weighting
suspicious edges \citep{jin2020prognn,zhang2020gnnguard}.  These methods remain
node-, edge-, or subgraph-level rather than modality-level selection methods.

MAGs introduce finer-grained noise because modalities of the same neighbor can
contribute unequally to a target.  Existing samplers select nodes, edges, or
subgraphs and cannot retain one modality while rejecting another from the same
node.  This motivates modality-level selection: topology constrains local
candidates, while each modality is evaluated independently.  Moreover, useful
modality evidence need not be confined to the local neighborhood, so semantic
compatibility should guide both local selection and complementary retrieval
beyond it.  The central decision is therefore which node--modality evidence
should enter a target-specific context.

\subsection{Transformers for Graph and Multimodal Learning}

Transformers are increasingly used for graph and multimodal representation
learning.

\textbf{Graph Transformers.}
Graph Transformers add structural bias through Positional Encodings
(PE), including Laplacian PE
\citep{DBLP:journals/corr/abs-2106-03893,DBLP:journals/corr/abs-2012-09699}
and Random Walk PE \citep{DBLP:journals/corr/abs-2110-07875}, or through
structure-biased attention, as in Graphormer
\citep{DBLP:journals/corr/abs-2106-05234}.

\textbf{Multimodal Transformers.}
Multimodal models use single-stream architectures such as VideoBERT
\citep{DBLP:journals/corr/abs-1904-01766} or dual streams such as ViLBERT
\citep{DBLP:journals/corr/abs-1908-02265}.  Mixture-of-Experts supports modality
specialization \citep{shazeer2017outrageouslylargeneuralnetworks}; MoMa
\citep{lin2024momaefficientearlyfusionpretraining}, for example, combines
modality-specific experts with shared attention.
However, their intersection remains underexplored, particularly how to model
fine-grained cross-modal interactions while retaining graph structural
dependencies inside a unified encoder.

%% file: sections/methods.tex
\section{Method}
\label{sec:method}

\subsection{Overview}
\label{sec:framework}

\begin{figure*}[t]
    \centering
    \includegraphics[width=\textwidth]{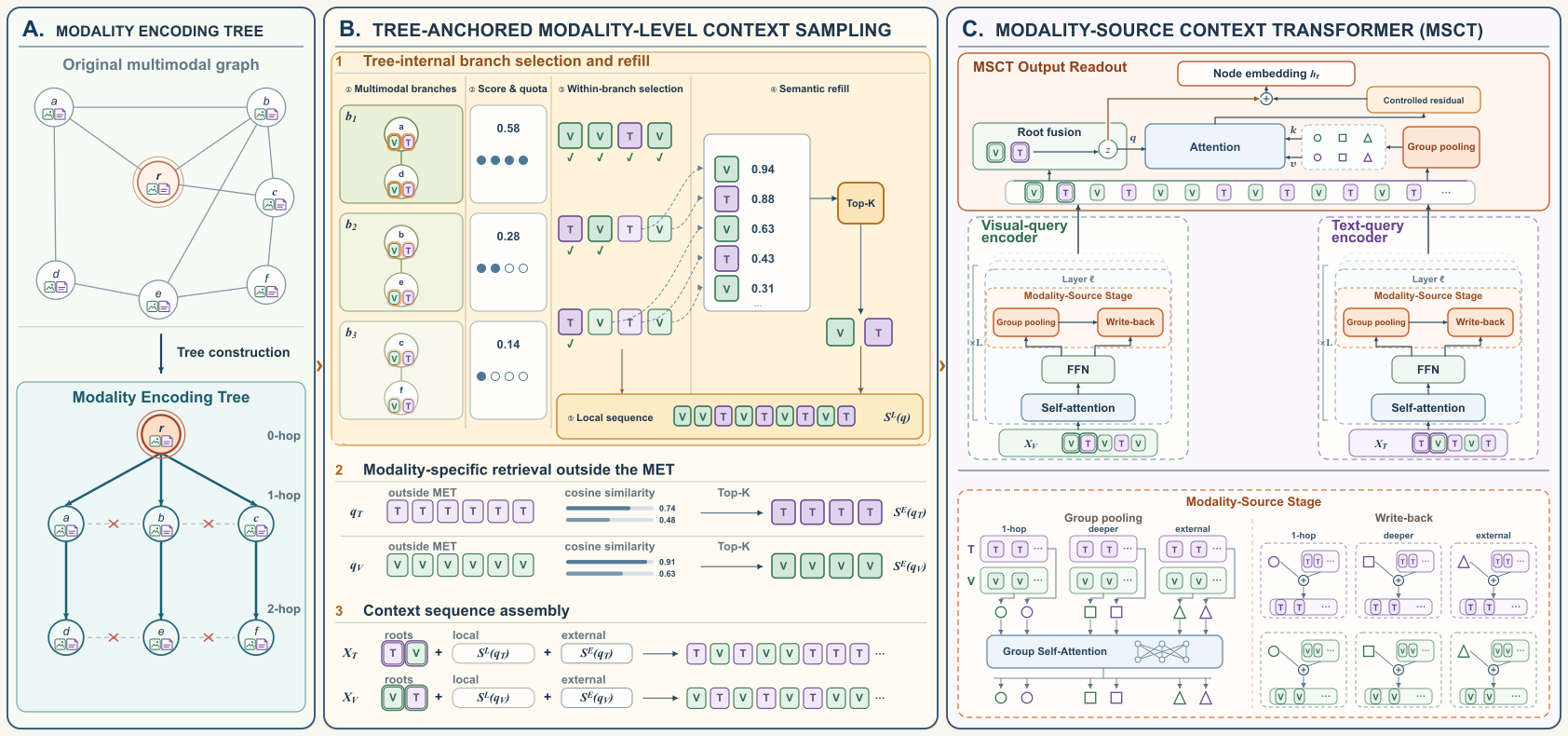}%
    \caption{TEAM framework: (A) MET construction; (B) tree-anchored
    modality-level context sampling; and (C) MSCT.  MSCT jointly performs
    modality-source-aware token interaction and group-conditioned target
    readout.}
    \label{fig:framework}
    \Description{TEAM first converts a local multimodal neighborhood into a
    rooted Modality Encoding Tree.  It selects local text and visual tokens by
    branch allocation and semantic refill, appends modality-specific tokens
    retrieved from outside the tree, and encodes two query sequences.  Within
    each query-specific token encoder, a unified modality--source interaction
    operation constructs six groups, exchanges information across them, and
    writes each refined group to the matching token cohort.  The MSCT output
    readout attends to the final group memory to form the node representation.}
\end{figure*}

\textbf{TEAM} combines tree-anchored sampling with cross-modal context
encoding.  As shown in Figure~\ref{fig:framework},
it unfolds the local neighborhood into a Modality Encoding Tree (MET), where
text and visual attributes form separate tokens with branch and hop provenance.
For each modality query, branch-aware allocation and semantic refill select
local tokens, while modality-specific retrieval appends compatible external
tokens.  The resulting sequences enter a Modality-Source Context Transformer
(MSCT), which coordinates six modality--source groups through matched token
write-back and a root-guided grouped readout.  The sampler therefore determines
which evidence enters each query, while MSCT models how evidence with different
modal and structural roles interacts.  Thus, topology organizes context while
semantics selects modality-level evidence.
These stages address complementary decisions: sampling controls which bounded
evidence is admitted, whereas MSCT controls how its modality and provenance
identities shape subsequent interaction.  The encoder therefore operates on a
selective, structured context rather than compensating for indiscriminate
neighborhood aggregation.

\subsection{Tree-Anchored Modality-Level Context Sampling}
\label{sec:evidence_acquisition}

\noindent\textit{Motivation.}
Node-level sampling retains every modality of a selected neighbor even when
only one is relevant.  A local neighborhood may also contain semantically weak
or modality-mismatched evidence, while structural proximity alone cannot
identify useful node--modality evidence.  We therefore organize candidates as a
first-visit tree with branch and hop provenance, rank its tokens by
target-conditioned relevance, and supplement the selected context with a small
modality-specific set from outside the MET.

\paragraph{Modality Encoding Tree.}
For target node $r$, TEAM converts its bounded local graph $\mathcal G_r$ into a
first-visit tree
\begin{equation}
    \mathcal T_r=\operatorname{BFSUnique}(\mathcal G_r,r;H,N_{\max}),
    \label{eq:met}
\end{equation}
where $H$ bounds the depth and $N_{\max}$ the number of retained nodes.  Each
non-root node $u$ receives one parent, hop index $h_u$, and first-hop branch
$b_u$; repeated appearances caused by cycles are removed.  Its text and visual
attributes are represented separately as $e_{u,T}$ and $e_{u,V}$.  The
non-root node set is $\mathcal V_r^L=\mathcal V(\mathcal T_r)\setminus\{r\}$;
expanding every $u\in\mathcal V_r^L$ by modality gives the local token set
$\mathcal C_r^L=\{e_{u,m}:u\in\mathcal V_r^L,m\in\{T,V\}\}$.

We construct a visual-query context ($q=V$) and a textual-query context
($q=T$).  Let $\psi_m$ map modality $m$ into a shared semantic space and define
\begin{equation}
    \mathbf z_i^m=\operatorname{Norm}\!\left(\psi_m(\mathbf x_i^m)\right),
\end{equation}
where $\operatorname{Norm}$ denotes $\ell_2$ normalization and $\psi_m$ is the
identity when the input features are already aligned.  For
$e_{u,m}\in\mathcal C_r^L$, we score each node--modality token by
\begin{equation}
\label{eq:token_score}
 a_q(e_{u,m})=
 \lambda_s(\mathbf z_r^q)^\top\mathbf z_u^m
 +\frac{\lambda_h}{h_u}+\lambda_d\widetilde d_{r,u},
\end{equation}
where $\lambda_s,\lambda_h,\lambda_d\geq0$ and $\widetilde d_{r,u}$ is the
log-degree of $u$ normalized within the local tree.  Semantic compatibility
provides the main ranking signal, while inverse hop and normalized degree retain
local structural priors.  We use $\operatorname{Top}_k(\mathcal C;a)$ to denote
up to $k$ available elements in $\mathcal C$ with the largest scores under $a$
for subsequent selection.

\paragraph{Branch-aware local refinement.}
Let $\mathcal C_{r,b}^{L}$ contain tokens sharing first-hop ancestor $b$.
Allocating more positions to a promising branch captures a broader semantic
view of that structural route.  We estimate its importance from leading scores,
\begin{equation}
 I_{q,b}=\operatorname{Mean}\!\left\{
 a_q(e):e\in\operatorname{Top}_{R}(\mathcal C_{r,b}^{L};a_q)
 \right\}.
 \label{eq:branch_importance}
\end{equation}
Here $R$ is the maximum number of leading tokens used to summarize a branch.
Let $K_L$ be the local context budget and $K_A\leq K_L$ the capacity
assigned through branch-aware allocation.  For the nonempty branches
$\mathcal B_r$, we form
\begin{equation}
 p_{q,b}=\frac{\exp(I_{q,b}/\tau)}
 {\sum_{b'\in\mathcal B_r}\exp(I_{q,b'}/\tau)},\qquad
 \boldsymbol\kappa_q=\operatorname{CapRound}
 (K_A\mathbf p_q;\mathbf c_q),
 \label{eq:branch_quota}
\end{equation}
where $\tau$ is the allocation temperature and $\mathbf c_q$ contains the
per-branch caps determined by candidate availability and the maximum branch
capacity.  Here $p_{q,b}$ is branch $b$'s normalized allocation share, so
$K_Ap_{q,b}$ is its real-valued proposed quota.  A lower $\tau$ concentrates
capacity on higher-importance branches, whereas a higher $\tau$ yields a more
even allocation.  $\operatorname{CapRound}$ is our deterministic capped-rounding
operator: it converts $K_A\mathbf p_q$ into integer quotas, enforces
$0\leq\kappa_{q,b}\leq c_{q,b}$, and redistributes residual capacity without
exceeding $K_A$.  Thus, $K_A$ controls the branch-planned portion of the total
local budget $K_L$.  The branch-allocated subset is
\begin{equation}
 \mathcal A_q=\bigcup_{b\in\mathcal B_r}
 \operatorname{Top}_{\kappa_{q,b}}(\mathcal C_{r,b}^{L};a_q).
\end{equation}
The highest-scoring unused tokens then refill the remaining local capacity:
\begin{equation}
 \mathcal S_q^{L}=\mathcal A_q\cup
 \operatorname{Top}_{K_L-|\mathcal A_q|}
 (\mathcal C_r^{L}\setminus\mathcal A_q;a_q).
 \label{eq:local_selection}
\end{equation}
Branch allocation gives semantically promising structural routes more context,
whereas refill assigns the remaining positions by token-level relevance over
all unused local candidates.  $\mathcal S_q^L$ is the final local token set
passed to the encoder.

\paragraph{Modality-specific retrieval outside the MET}
Let $\mathcal B_r^E$ be the external candidate bank formed by accessible graph
nodes outside the target's MET, and let $K_E$ be the
external context budget.  With
$s_q^E(r,j)=(\mathbf z_r^q)^\top\mathbf z_j^q$, text and vision
retrieve independently:
\begin{equation}
\begin{aligned}
 \mathcal R_q^E&=\operatorname{Top}_{K_E}
 (\mathcal B_r^E;s_q^E),\\
 \mathcal S_q^E&=\{e_{j,q}:j\in\mathcal R_q^E\}.
\end{aligned}
\label{eq:external_selection}
\end{equation}
The two queries may retrieve different nodes, and each retrieved node contributes
only the modality matched to that query in the corresponding context sequence.

\paragraph{Context assembly.}
Let $\bar q$ denote the other modality.  Concatenation, denoted by $\Vert$,
produces
\begin{equation}
 \mathbf X_q=[e_{r,q},e_{r,\bar q}]
 \Vert\mathcal S_q^L
 \Vert\mathcal S_q^E,\qquad q\in\{T,V\}.
 \label{eq:assembled_sequences}
\end{equation}
Tokens within each segment follow their selection order and carry their
modality, root/context role, hop, and source type; branch identity guides local
allocation before assembly.  Padding and masking
give a common encoder width.  Together, the ordered local and external segments
form a query-specific context while preserving source provenance for encoding.

\subsection{Modality-Source Context Transformer}
\label{sec:msct}

\noindent\textit{Motivation.}
The sampler specifies which text and visual evidence enters each context, but
not how these tokens should interact.  Self-attention lets all selected tokens
exchange content, but treats them as one mixed sequence rather than explicitly
aggregating and routing evidence by modality and source.  MSCT therefore
integrates token interaction and modality--source interaction in each encoding
layer, and reuses the same evidence identities for target-conditioned output
readout.

\noindent\textit{MSCT layer.}
The two query sequences use separate $L$-layer branches.  Each MSCT layer
combines a token-interaction stage with a subsequent modality--source stage.
With $\widehat{\mathbf H}_q^{(0)}=\mathbf H_q^{(0)}$, the token stage applies
standard self-attention and feed-forward transformation:
\begin{equation}
 \mathbf H_q^{(\ell)}=
 \operatorname{TransformerLayer}_{q,\ell}
 (\widehat{\mathbf H}_q^{(\ell-1)}),\qquad 1\leq\ell\leq L.
 \label{eq:token_layer}
\end{equation}
The modality--source stage begins with \emph{Group Pooling} of the resulting
valid non-root states according to
\begin{equation}
 \Gamma=\{(m,p):m\in\{T,V\},\ p\in\{L_1,L_{2+},E\}\},
 \label{eq:six_slots}
\end{equation}
where $L_1$, $L_{2+}$, and $E$ denote one-hop local, multi-hop local, and
external sources.  If $\mathcal I_{q,g}$ indexes the valid tokens in group $g$,
its state is their mean:
\begin{equation}
 \mathbf s_{q,g}^{(\ell)}=
 \frac{1}{|\mathcal I_{q,g}|}
 \sum_{i\in\mathcal I_{q,g}}\mathbf h_{q,i}^{(\ell)}.
 \label{eq:slot_pool}
\end{equation}
Padding tokens are excluded from $\mathcal I_{q,g}$, and empty groups are
masked.  The pooled group states then undergo \emph{Group Self-Attention}, with
learned group identities added before attention, to obtain
$\widetilde{\mathbf s}_{q,g}^{(\ell)}$.

For group $g=(m,p)$, let $\bar m$ denote the other modality.  MSCT compares the
group state with the other-modality root $\mathbf r_{q,\bar m}^{(\ell)}$:
\begin{equation}
 \mathbf d_{q,g}^{(\ell)}=\mathbf W_u\operatorname{GELU}
 \!\left(\mathbf W_d\!\left[
 \widetilde{\mathbf s}_{q,g}^{(\ell)}-
 \operatorname{LN}(\mathbf r_{q,\bar m}^{(\ell)})\right]\right).
 \label{eq:crossmodal_innovation}
\end{equation}
The difference emphasizes information not already expressed by the other
modality.  Let $c_{q,g}^{(\ell)}=\cos(\mathbf s_{q,g}^{(\ell)},
\mathbf r_{q,\bar m}^{(\ell)})$ and
$\rho_{q,g}=|\mathcal I_{q,g}|/\sum_{g'}|\mathcal I_{q,g'}|$.  We collect the
gate inputs as $\boldsymbol\xi_{q,g}^{(\ell)}=[c_{q,g}^{(\ell)},\rho_{q,g},
\mathbb I[p=E]]$ and compute
\begin{equation}
 \gamma_{q,g}^{(\ell)}=\gamma_{\max}\tanh\phi(\boldsymbol\xi_{q,g}^{(\ell)}),
 \qquad \mathbf u_{q,g}^{(\ell)}=\gamma_{q,g}^{(\ell)}\mathbf d_{q,g}^{(\ell)}.
 \label{eq:group_gate}
\end{equation}
Here $\phi$ is a learned gating network and $\gamma_{\max}$ bounds the update.
The update $\mathbf u_{q,g}^{(\ell)}$ is written to the FFN states of tokens
in the matching group $g$.  This matched residual routing returns group-level
evidence to its original token cohort while retaining the modality--source
partition and sequence order.  The
token stage, group interaction, and matched write-back together form one MSCT
layer:
\begin{equation}
\begin{aligned}
 \widehat{\mathbf H}_q^{(\ell)}
 &=\operatorname{WriteBack}\!\left(
 \mathbf H_q^{(\ell)},\{\mathbf u_{q,g}^{(\ell)}\}_{g\in\Gamma}\right)\\
 &\equiv \operatorname{MSCTLayer}_{q,\ell}
 (\widehat{\mathbf H}_q^{(\ell-1)},\Gamma),
 \qquad 1\leq\ell\leq L.
\end{aligned}
 \label{eq:grouped_context}
\end{equation}
The query branches keep separate token states and Transformer parameters; the
modality--source interaction parameters are shared across the two branches and
reused in every MSCT layer.

\paragraph{MSCT output readout.}
Let $\mathbf h_V$ and $\mathbf h_T$ be the final visual and textual root states
from $\widehat{\mathbf H}_V^{(L)}$ and $\widehat{\mathbf H}_T^{(L)}$.  Their
fused target state is
\begin{equation}
 \mathbf z=\operatorname{LN}
 \left(\mathbf W_o[\mathbf h_V\Vert\mathbf h_T]+\mathbf b_o\right).
 \label{eq:root_anchor}
\end{equation}
Root fusion summarizes the target, whereas the six final group states retain
source-specific context.  MSCT pools valid non-root states from both branches
into memory $\mathbf M=[\mathbf m_g]_{g\in\Gamma}$ using the same group
identities.  Let $s_g$ denote support normalized across groups and
$\Gamma^+$ the groups with nonzero support.  With
$\mathbf k_g=\mathbf W_k\mathbf m_g+\mathbf e_g$ and
$\mathbf v_g=\mathbf W_v\mathbf m_g+\mathbf e_g$, the target reads valid memory
cells as
\begin{equation}
\begin{split}
 \ell_g&=\frac{Q(\mathbf z)\mathbf k_g^\top}{\sqrt{d_a}}
 +\log(s_g+\epsilon),\\
 \alpha_g&=\operatorname{softmax}_{g\in\Gamma^+}(\ell_g),\qquad
 \mathbf c=\sum_{g\in\Gamma^+}\alpha_g\mathbf v_g.
\end{split}
 \label{eq:slot_readout}
\end{equation}
This gives a target-conditioned path to the grouped context.  To keep that path
corrective rather than dominant, MSCT uses a mean-zero, RMS-capped residual:
\begin{equation}
\begin{split}
 \mathbf u&=\eta\tanh(\mathbf W_c\mathbf c),\qquad
 \bar{\mathbf u}=\mathbf u-
 \frac{\mathbf 1^\top\mathbf u}{d}\mathbf 1,\\
 \boldsymbol\delta&=\bar{\mathbf u}\min\!\left\{1,
 \frac{\eta}{\sqrt{\|\bar{\mathbf u}\|_2^2/d+\epsilon^2}}\right\},
 \qquad
 \mathbf h_r=\mathbf z+\boldsymbol\delta.
\end{split}
\label{eq:node_readout}
\end{equation}
The residual has zero feature mean and RMS at most $\eta$; it adjusts the node
state without replacing the fused root state.

\subsection{Task Adaptation}
\label{sec:task_adaptation}

TEAM returns a node representation $\mathbf h_r$ that is shared across task
interfaces.  For node classification, a linear classifier maps $\mathbf h_r$
to class logits and is optimized by cross-entropy.  For link prediction, an MLP
predictor maps the element-wise product of two endpoint representations to an
edge probability and is optimized on positive and negative edges.  For G2Text
and G2Image, a task-specific projection maps $\mathbf h_r$ into the conditioning
space of the language or diffusion backbone, respectively, and training follows
the native generation objective of the corresponding task.

\subsection{Theoretical Analysis of Tree-Anchored Semantic Sampling}
\label{sec:theory}

We provide encoder-independent theoretical support for TEAM's sampler by
analyzing its successive stages: branch allocation, quota-constrained local
selection, and equal-budget external supplementation.

\begin{lemma}[Entropy-regularized branch allocation]
\label{lem:branch_allocation}
Let $\Delta_r=\{\mathbf p\geq\mathbf 0:\sum_b p_b=1\}$ and
$\mathcal H(\mathbf p)=-\sum_b p_b\log p_b$.  Before capping and rounding, the
shares in Eq.~\eqref{eq:branch_quota} uniquely solve
\begin{equation}
 \mathbf p_q=\arg\max_{\mathbf p\in\Delta_r}
 \left\{\sum_{b\in\mathcal B_r}p_b I_{q,b}
 +\tau\mathcal H(\mathbf p)\right\}.
 \label{eq:branch_regularized_objective}
\end{equation}
Thus, higher importance yields no smaller proposed share, while $\tau$ balances
concentration and branch coverage.
\end{lemma}

\emph{Proof.}
Strict concavity gives uniqueness.  The Lagrange condition yields
$p_b\propto\exp(I_{q,b}/\tau)$; normalization gives Eq.~\eqref{eq:branch_quota}
and the stated monotonicity.

\begin{lemma}[Score-optimal local selection under branch quotas]
\label{lem:local_optimality}
Assume $\{\mathcal C_{r,b}^L\}_b$ partitions $\mathcal C_r^L$, and fix feasible
integer quotas with $\sum_b\kappa_{q,b}\leq K_L$.  The branch selection followed
by Eq.~\eqref{eq:local_selection} satisfies
\begin{equation}
 \mathcal S_q^L\in\arg\max_{\substack{\mathcal S\subseteq\mathcal C_r^L,
 |\mathcal S|=K_L\\
 |\mathcal S\cap\mathcal C_{r,b}^L|\geq\kappa_{q,b}\ \forall b}}
 \sum_{e\in\mathcal S}a_q(e).
 \label{eq:quota_optimal_selection}
\end{equation}
\end{lemma}

\emph{Proof.}
Replacing a lower-scoring same-branch token by a missing top-$\kappa_{q,b}$
token preserves feasibility, so an optimum contains $\mathcal A_q$.  The global
Top-$k$ refill then maximizes the remaining contribution.

\begin{theorem}[Equal-budget benefit of external semantic supplementation]
\label{thm:semantic_augmentation}
Let $U_q(e)$ be additive latent utility and
$\mathcal C_q^{L+}=\mathcal C_r^L\setminus\mathcal S_q^L$.  For a candidate pool
$\mathcal C$, define
$U_q^{(k)}(\mathcal C)=\max_{|\mathcal A|=k,\mathcal A\subseteq\mathcal C}
\sum_{e\in\mathcal A}U_q(e)$.  Suppose a nondecreasing calibration $g_q$
satisfies
\begin{equation}
 \left|g_q(s_q^E(r,e))-U_q(e)\right|\leq\epsilon_q,
 \qquad e\in\mathcal B_r^E.
 \label{eq:score_utility_error}
\end{equation}
If both pools contain at least $K_E$ candidates and
\begin{equation}
 U_q^{(K_E)}(\mathcal B_r^E)-U_q^{(K_E)}(\mathcal C_q^{L+})
 >2K_E\epsilon_q,
 \label{eq:external_margin_condition}
\end{equation}
then TEAM's retrieved set has greater utility than any equal-size continuation
from unused local candidates:
\begin{equation}
 \sum_{e\in\mathcal S_q^L\cup\mathcal S_q^E}U_q(e)>
 \sum_{e\in\mathcal S_q^L}U_q(e)+U_q^{(K_E)}(\mathcal C_q^{L+}).
 \label{eq:equal_budget_external_gain}
\end{equation}
\end{theorem}

\emph{Proof.}
Monotonicity makes semantic Top-$K_E$ optimal for calibrated utility, and
Eq.~\eqref{eq:score_utility_error} bounds its oracle regret by
$2K_E\epsilon_q$.  Thus Eq.~\eqref{eq:external_margin_condition}, after adding
the common $\mathcal S_q^L$, yields Eq.~\eqref{eq:equal_budget_external_gain}.

%% file: sections/experiments.tex
\setlength{\dbltextfloatsep}{8pt plus 2pt minus 2pt}
\setlength{\textfloatsep}{8pt plus 2pt minus 2pt}
\setcounter{dbltopnumber}{1}

\input{sections/layout_experiment_plates.tex}

\section{Experiments}
\label{sec:experiments}

We organize the evaluation around six research questions:
\noindent\textbf{RQ1:} How effective is TEAM on graph learning tasks?
\textbf{RQ2:} How effective is TEAM on multimodal tasks?
\textbf{RQ3:} How do the main sampling and Transformer components affect TEAM?
\textbf{RQ4:} How robust is TEAM to structural, modal, and supervisory noise?
\textbf{RQ5:} How sensitive is TEAM to its local and external evidence budgets?
\textbf{RQ6:} What computational overhead does TEAM introduce?

\subsection{Experimental Setup}
\label{subsec:setup}

\paragraph{Datasets and tasks.}
Following OpenMAG~\cite{wan2026openmag}, \emph{graph-learning tasks} include
node classification on Movies, Toys, Grocery, and RedditS and link prediction
on TN, Bili-Movie, KU, and QB, while \emph{multimodal tasks} include G2Text on Flickr30k
\cite{plummer2015flickr30k} and G2Image on SemArt
\cite{garcia2018read}.  Graph prediction uses its features, splits, and evaluators;
generation follows each benchmark's setup and evaluator.  We report Accuracy/Macro-F1 for node classification, MRR/Hits@3 for
link prediction, BLEU-4/ROUGE-L/CIDEr for G2Text, and CLIP-Score/DINOv2-Score
for G2Image.

\paragraph{Baselines.}
Our baselines span four model families.  Graph neural networks include
GCN~\cite{kipf2017semi}, GAT~\cite{velickovic2018graph}, and
GraphSAGE~\cite{hamilton2017inductive}, while graph Transformers include
Graphormer~\cite{DBLP:journals/corr/abs-2106-05234} and
DUALFormer~\cite{zhuo2025dualformer}.  Multimodal Transformers include
MMBT~\cite{kiela2019mmbt} and VLMo~\cite{bao2022vlmo}, while multimodal graph
learners include MMGCN~\cite{wei2019mmgcn}, DGF~\cite{zheng2025dgf},
DMGC~\cite{guo2025disentanglinghomophilyheterophilymultimodal}, and
LGMRec~\cite{guo2024lgmrec}.  We also include a feature-only MLP.  Generation
comparisons share the same downstream decoder and evaluator, thereby
attributing differences to graph-encoder quality rather than to independently
designed task-specific generation pipelines and decoder choices.

\paragraph{Implementation.}
Unless varied, TEAM uses a two-hop local tree, $(K_L,K_E)=(16,4)$, two
384-dimensional encoder layers, eight attention heads, and dropout 0.5.  For
node classification, OpenMAG baselines use the prescribed learning rate of
$5\times10^{-3}$, while Transformer-based models use $10^{-4}$; training uses
batch size 512, at most 1,000 epochs, and patience 50.  For link prediction,
OpenMAG baselines use $10^{-3}$, while Transformer-based models use
$2\times10^{-4}$, with batch size 2,048 and at most 2,000 epochs.  Both graph tasks
sample 15 neighbors at each hop for all compared methods.  For
G2Text, we use a frozen OPT-125M language model and 768-dimensional
CLIP-ViT-L/14 features.  We train all
compared models with learning rate $2\times10^{-3}$ for 10 epochs, physical
batch size 4, gradient accumulation 8, and 200 warmup steps.  For G2Image, we
use Stable Diffusion v1.5 at $256\times256$ resolution.  We use learning rate
$5\times10^{-6}$, batch size 16, 10 epochs, and 1,000 training samples.  For
both generation tasks, target outputs are withheld from encoder inputs under
the shared benchmark graph construction.

\paragraph{Protocol and reporting.}
All methods use identical inputs, splits, and task evaluators within each
comparison; generation models additionally share the downstream backbone and
evaluation pipeline.  All experiments are repeated three times and summarized
by their mean and standard deviation.
For link prediction, edge splitting, negative sampling, and ranking evaluation
follow the original OpenMAG protocol.
Bold and underlined values denote the best and second-best results, respectively.

\subsection{Performance on Graph-Centric Tasks}
\label{subsec:graph_tasks}

\subsubsection{Node Classification}

Table~\ref{tab:nc_main} shows that TEAM attains the highest mean Accuracy and
Macro-F1 on all four graphs.  On Movies, the modest Accuracy lead is accompanied
by a clearer Macro-F1 margin, indicating that the improvement extends beyond
majority-class decisions.  Toys and Grocery show broader gains on both metrics,
suggesting that structured multimodal context is especially useful when
substantial headroom remains.  RedditS further improves near an already high
performance ceiling.  Consistent gains across these class distributions and
performance regimes show that TEAM is not tied to one metric or baseline range,
and remains effective as dataset scale and task difficulty vary.

\vspace{0.30em}
\subsubsection{Link Prediction}

TEAM also ranks first in MRR and Hits@3 on all four link-prediction datasets in
Table~\ref{tab:lp_main}.  These metrics provide complementary views of the same
ranking: MRR rewards placing the correct endpoint as early as possible, whereas
Hits@3 measures whether it appears among the first three candidates.  TEAM's
joint lead therefore shows consistent improvement in both ranking position and
top-three retrieval.  TN is the tightest comparison, particularly on Hits@3,
yet TEAM remains best on both metrics; the clearer margins on Bili-Movie, KU,
and QB show that the advantage persists across different candidate spaces.
Agreement across both metrics also reduces the possibility that the result is
driven only by one favorable cutoff or a few isolated first-ranked predictions.
Together with node classification, these results show that TEAM remains
effective for both node-wise prediction and pairwise link ranking through
task-specific prediction heads.

\subsection{Performance on Multimodal Tasks}
\label{subsec:modality_tasks}

Table~\ref{tab:generation_main} evaluates TEAM on text and image generation.
Both benchmarks construct graph neighborhoods from pretrained semantic
features, so all methods begin with a strong relevance prior.  TEAM
nevertheless leads all G2Text baselines on BLEU-4, ROUGE-L, and CIDEr,
showing that the complete TEAM encoder can exploit graph context beyond relying
on construction-time similarity alone.  On G2Image, TEAM attains the highest mean
on both CLIP-Score and DINOv2-Score, supporting improvements in semantic
alignment and visual representation quality.  Although G2Text and G2Image use
different downstream decoders, TEAM provides both with a common graph
representation through task-specific output projections; its sampling and
context-interaction design remain unchanged.  The gains also span complementary
evaluation criteria.  BLEU-4 and ROUGE-L emphasize lexical overlap, whereas
CIDEr emphasizes agreement with reference descriptions; their joint increase
shows that the G2Text result is not specific to one text metric.  Likewise,
CLIP-Score measures text--image alignment and DINOv2-Score reflects visual
representation similarity, so leading on both supports semantic consistency
without sacrificing visual fidelity.  Taken together, these results show that
TEAM remains effective across text and image generation even when the benchmark
topology already encodes a strong semantic relevance prior at construction time.

\subsection{Ablation Study}
\label{subsec:ablation}

Figure~\ref{fig:ablation} evaluates three targeted controls.  The
\emph{Randomized Sampler} retains the same tree construction and matched
local--external capacities, but replaces semantic selection in both local
refinement and external retrieval with random sampling.  It therefore isolates
the value of relevance without changing the amount of available context.  The
\emph{Naive Sampler} applies local semantic
top-$k$ without branch-aware allocation or refill, testing the proposed
procedure beyond relevance alone.  The \emph{Naive Transformer} replaces MSCT
with a standard Transformer and root concatenation, directly testing our
proposed Transformer under the same selected context.  Full TEAM leads in all four
cases.  The sampler controls have larger effects on Toys, Grocery, and QB,
whereas the Transformer control is especially clear on TN.  These distinct
gaps support complementary contributions from semantic relevance, tree-aware
organization, and modality--source-aware interaction.  Full TEAM and the
Transformer control contain 8.73M and 8.63M parameters; their 1.14\% difference
is too small to explain the gain by capacity alone.

The controls are complementary: randomization removes relevance while retaining
the tree and matched capacities; the naive sampler retains relevance but removes
branch budgeting and refill; and the Transformer control receives the same
selected context but uses a standard Transformer encoder.  Their different gap
sizes reflect dataset-specific bottlenecks, whereas their consistent direction
shows that gains persist across tasks and cannot be reduced to one isolated
architectural change within TEAM.

\FloatBarrier

%% file: sections/layout_experiment_plates.tex

\begin{table*}[t!]
\input{sections/table_nc.tex}
\vspace{0.8em}
\input{sections/table_lp.tex}
\Description{Two tables report node-classification and link-prediction results
across four multimodal graph benchmarks, comparing TEAM with graph,
Transformer, and multimodal baselines.}
\end{table*}

\begin{figure*}[t!]
\input{sections/table_multimodal.tex}

\vspace{0.15em}
\begin{minipage}{\textwidth}
\centering
\includegraphics[width=0.90\textwidth]{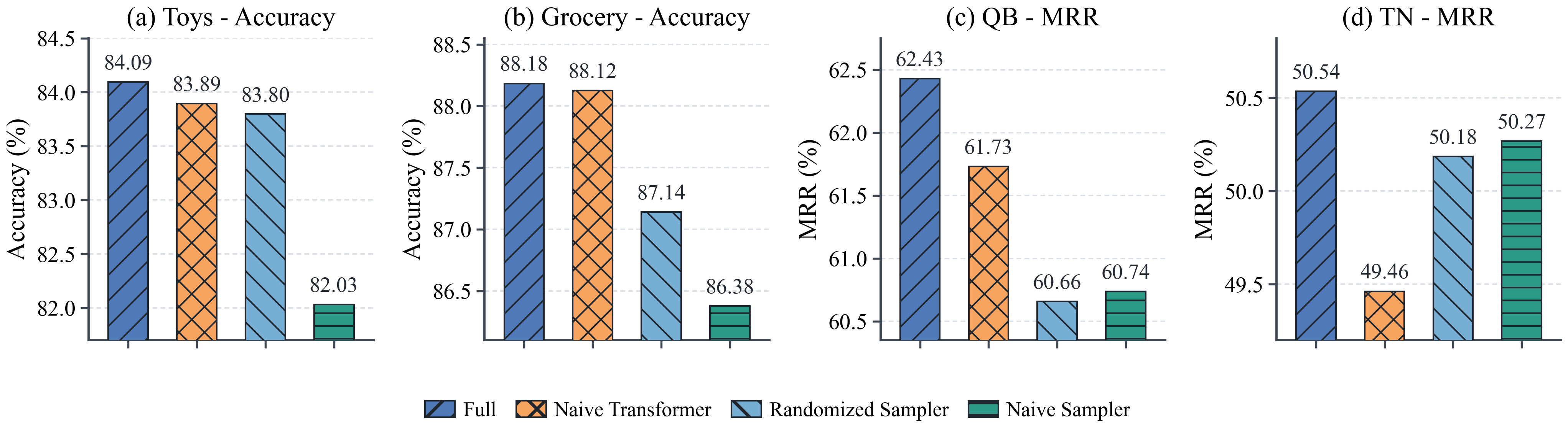}
\captionof{figure}{Component ablation on node classification and link prediction.}
\label{fig:ablation}
\Description{Four bar charts comparing Full TEAM with Naive Transformer,
Randomized Sampler, and Naive Sampler controls.}
\end{minipage}
\end{figure*}

\begin{figure*}[t]
\begin{minipage}{\textwidth}
\centering
\begin{minipage}[b]{0.49\textwidth}
\centering
\includegraphics[width=\linewidth]{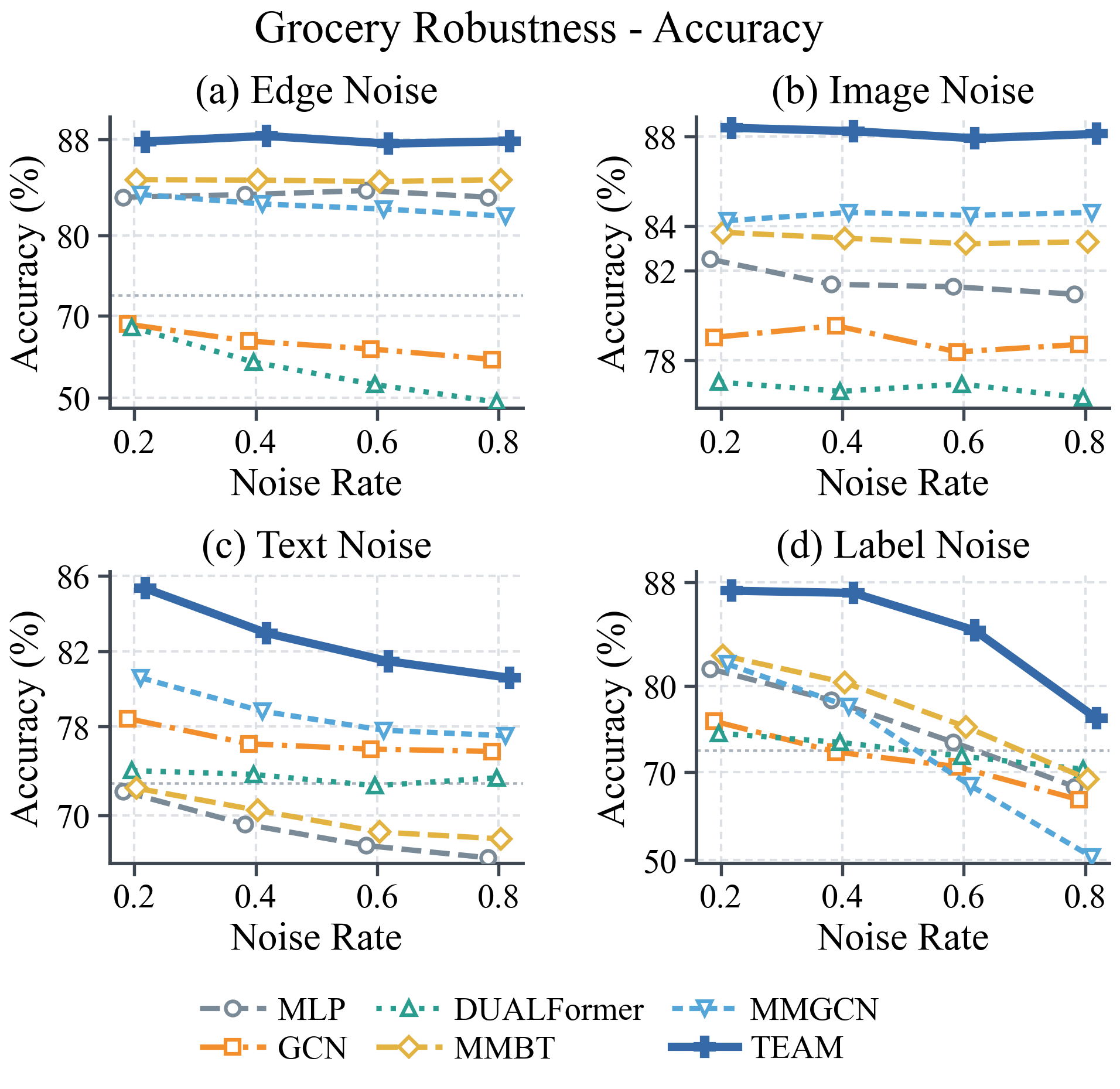}
\end{minipage}
\hfill
\begin{minipage}[b]{0.49\textwidth}
\centering
\includegraphics[width=\linewidth]{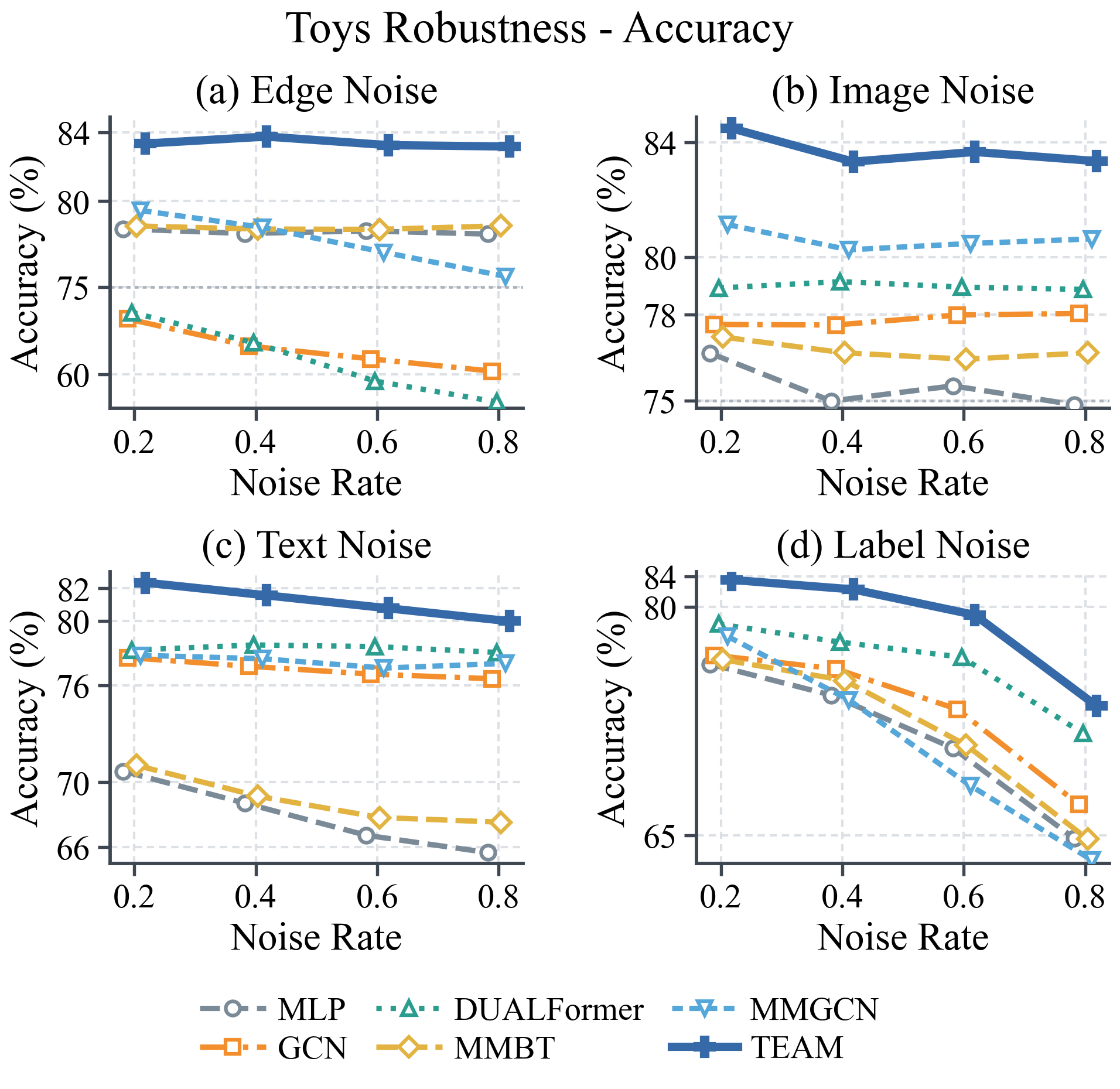}
\end{minipage}
\captionof{figure}{Robustness to edge, image, text, and label noise on Grocery
and Toys.  Both panels report Accuracy (\%).}
\label{fig:robustness}
\Description{Two four-panel plots comparing TEAM and five baselines under
increasing edge, image, text, and label noise.  Both panels report Accuracy.}
\end{minipage}
\end{figure*}

\begin{figure}[t]
\centering
\includegraphics[width=\columnwidth]{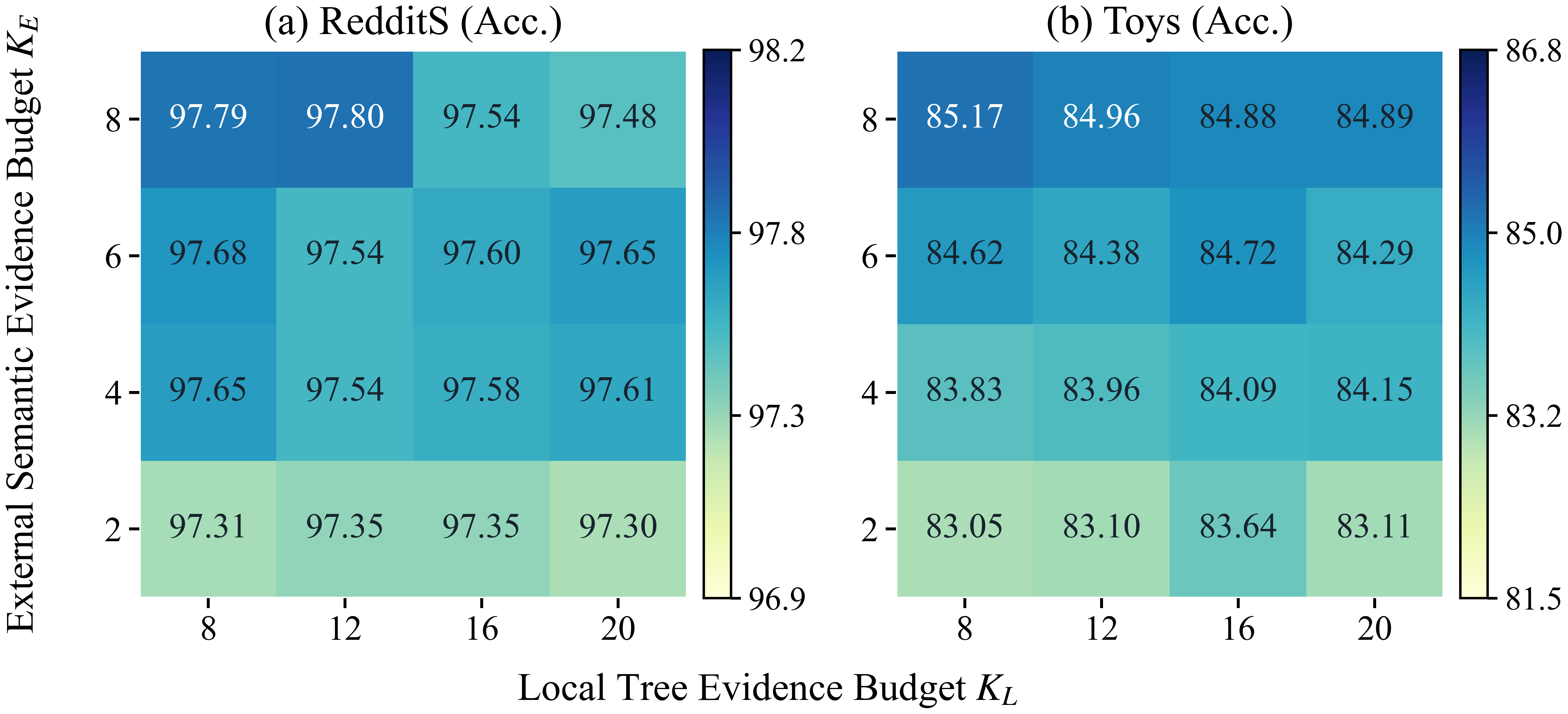}
\captionof{figure}{Sensitivity to local and external evidence budgets.}
\label{fig:sensitivity}
\Description{Two four-by-four heatmaps report RedditS and Toys test Accuracy as the
local and external evidence budgets vary.}
\end{figure}

\begin{figure*}[t!]
\input{sections/tables_efficiency.tex}
\vspace{1.55em}
\begin{minipage}[t]{0.49\textwidth}
\vspace{0pt}
\input{sections/table_efficiency_measured.tex}
\vspace{0.80em}

\noindent{\large\bfseries 4.5\quad Robustness Analysis}\par
\vspace{0.05em}
Figure~\ref{fig:robustness} evaluates structural, modal, and supervisory
corruption on Grocery and Toys.  Edge noise inserts uniformly sampled edges;
image and text noise permute sampled-node features; and label noise randomizes
only training labels.  Each setting is retrained after perturbations are applied
before MET construction, semantic scoring, and retrieval, so the corrupted
source supplies no clean signal.  TEAM remains strongest at the largest tested
rate: edge and image noise cause limited variation, text noise yields a gradual
decline, and label noise is more damaging because it directly corrupts
supervision.  The trends indicate distributed use of graph and multimodal
context rather than reliance on one fragile source.

\vspace{0.85em}

\noindent{\large\bfseries 4.6\quad Hyperparameter Sensitivity}\par
\vspace{0.05em}
Figure~\ref{fig:sensitivity} varies $K_L\in\{8,12,16,20\}$ and
$K_E\in\{2,4,6,8\}$ on RedditS and Toys.  Both datasets benefit more
consistently from a larger external budget than from a larger local budget,
reflecting the value of semantic evidence outside the local tree.  Toys improves
markedly with $K_E$, while its response to $K_L$ becomes non-monotonic once
external context is sufficient.  RedditS varies less near saturation, but larger
external budgets still form the strongest region and neighboring settings remain comparable.
The same overall trend appears across both datasets, but at different magnitudes: Toys
retains substantial headroom, whereas RedditS shows smaller gains near its
performance ceiling.  Useful performance is spread over neighboring budget
pairs, supporting complementary local and external contexts rather than one
fragile configuration on either evaluated dataset.
\end{minipage}
\hfill
\begin{minipage}[t]{0.48\textwidth}
\vspace{0.85em}
\noindent{\large\bfseries 4.7\quad Efficiency Analysis}\par
\vspace{0.05em}
Tables~\ref{tab:complexity} and~\ref{tab:efficiency} relate TEAM's bounded
context design to its realized cost.  Table~\ref{tab:complexity} uses a common
full-graph abstraction for all methods.  Symbols $d_0,d,L,C,S,h$, and $P_\theta$
denote input/hidden widths, layers, maximum local candidates per target,
sequence length, heads, and parameters.  TEAM uses $\mathcal{O}(|V|C)$ local
construction and $\mathcal{O}(|V|^2d_0)$ exact semantic comparison before
bounded-sequence encoding.  Its space row is a conservative full-graph upper
bound; the implementation evaluates target nodes in mini-batches, while
Table~\ref{tab:efficiency} reports the realized peak cost.
Experiments run on a server with two NVIDIA RTX PRO 6000 Blackwell GPUs,
PyTorch 2.8, and CUDA 12.8.  Efficiency measurements use one GPU at a time
under the same batch size of 512.  Epoch time is averaged over ten training epochs and includes
TEAM's context construction, retrieval, forward pass, backpropagation, and
parameter update; one-time pretrained feature extraction is excluded.
On Movies, TEAM reserves 1,626 MiB of GPU memory, below most structure-aware and
Transformer baselines, while its 7.53-second epoch remains markedly faster than
Graphormer and VLMo.  Although lightweight GNNs remain faster, TEAM delivers the
strongest graph-learning results reported above with bounded per-target
interaction, yielding a favorable accuracy--efficiency trade-off without global
graph self-attention.

\vspace{\dimexpr2.00em+0.0\baselineskip\relax}
\noindent{\Large\bfseries 5\quad Conclusion}\par
\vspace{0pt}
We presented TEAM, which combines a Modality Encoding Tree, semantic
supplementation, and MSCT to organize multimodal graph context.  Branch-aware
refinement prioritizes neighborhood evidence, modality-specific retrieval
extends semantic coverage, and MSCT preserves modality and source identities
during interaction.  Results across graph learning and graph-conditioned
multimodal generation, together with ablation, robustness, sensitivity, and
efficiency analyses, support the sampling and encoding design.  Future work may
extend the external candidate bank while retaining bounded context and cost.
The unified encoder also supports graph prediction and multimodal generation
through a common interface.
More broadly, the results suggest that structural organization and semantic
selection are most useful as complementary operations rather than competing
alternatives.  This perspective provides a practical basis for adapting TEAM
to further multimodal graph domains while preserving target-centered
computation and bounded evidence access for each target.
\end{minipage}
\end{figure*}

%% file: sections/table_nc.tex
\begin{minipage}{\textwidth}
\centering
\captionof{table}{Node classification on four MAG benchmarks (\%). Best results
are bold and second-best results are underlined.}
\label{tab:nc_main}
\fontsize{7.2}{7.8}\selectfont
\setlength{\tabcolsep}{2.8pt}
\renewcommand{\arraystretch}{0.70}
\resizebox{0.94\textwidth}{!}{%
\begin{tabular}{l cc cc cc cc}
\toprule
\multirow{2}{*}{Method}
& \multicolumn{2}{c}{Movies}
& \multicolumn{2}{c}{Toys}
& \multicolumn{2}{c}{Grocery}
& \multicolumn{2}{c}{RedditS} \\
\cmidrule(lr){2-3}\cmidrule(lr){4-5}\cmidrule(lr){6-7}\cmidrule(lr){8-9}
& Acc & F1 & Acc & F1 & Acc & F1 & Acc & F1 \\
\midrule
MLP & 51.69$\pm$0.48 & 41.37$\pm$1.74 & 77.87$\pm$0.39 & 75.63$\pm$0.76 & 83.07$\pm$0.66 & 75.85$\pm$0.70 & 93.96$\pm$0.32 & 88.60$\pm$0.58 \\
\midrule
GCN & 52.57$\pm$0.34 & 43.88$\pm$0.15 & 77.64$\pm$0.12 & 73.97$\pm$0.49 & 78.65$\pm$0.48 & 69.33$\pm$0.68 & 94.42$\pm$0.27 & 90.41$\pm$0.47 \\
GAT & 50.86$\pm$0.47 & 43.55$\pm$1.54 & 78.55$\pm$0.24 & 74.70$\pm$0.60 & 80.04$\pm$0.58 & 70.10$\pm$0.74 & 94.07$\pm$0.23 & 90.20$\pm$0.33 \\
GraphSAGE & 53.47$\pm$0.19 & 44.11$\pm$1.91 & 79.10$\pm$0.81 & 75.60$\pm$1.28 & 82.14$\pm$0.74 & 74.37$\pm$0.29 & 95.96$\pm$0.04 & 91.80$\pm$0.42 \\
\midrule
Graphormer & 50.31$\pm$0.31 & 37.49$\pm$2.30 & 76.49$\pm$0.45 & 73.97$\pm$0.19 & 82.21$\pm$0.32 & 75.21$\pm$0.56 & 94.05$\pm$0.06 & 88.81$\pm$0.17 \\
DUALFormer & 52.09$\pm$0.38 & 41.68$\pm$0.91 & 78.97$\pm$0.12 & 75.06$\pm$0.11 & 77.12$\pm$0.41 & 64.20$\pm$0.51 & 94.39$\pm$0.28 & 88.23$\pm$0.63 \\
\midrule
MMBT & 51.33$\pm$0.67 & 37.94$\pm$1.59 & 77.82$\pm$0.27 & 75.36$\pm$0.27 & 83.87$\pm$0.97 & 77.64$\pm$1.13 & 94.06$\pm$0.19 & 87.81$\pm$0.11 \\
VLMo & 51.80$\pm$0.06 & 43.14$\pm$0.15 & 78.51$\pm$0.30 & 76.13$\pm$0.44 & 84.68$\pm$0.23 & 77.63$\pm$0.28 & 94.34$\pm$0.05 & 88.84$\pm$0.11 \\
\midrule
MMGCN & 53.12$\pm$0.31 & 45.09$\pm$0.81 & \underline{80.72$\pm$0.12} & \underline{78.54$\pm$0.25} & 84.53$\pm$0.60 & 77.07$\pm$0.87 & 96.66$\pm$0.13 & 92.53$\pm$0.61 \\
DGF & 52.16$\pm$0.09 & 43.27$\pm$1.76 & 80.52$\pm$0.26 & 77.17$\pm$0.45 & \underline{85.05$\pm$0.34} & \underline{77.68$\pm$0.70} & \underline{96.75$\pm$0.10} & \underline{92.63$\pm$0.08} \\
DMGC & 51.76$\pm$0.41 & 44.74$\pm$0.84 & 80.05$\pm$0.16 & 77.41$\pm$0.48 & 79.43$\pm$1.42 & 69.09$\pm$1.01 & 94.29$\pm$1.48 & 89.30$\pm$1.64 \\
LGMRec & \underline{53.79$\pm$0.42} & \underline{46.92$\pm$0.20} & 80.51$\pm$0.39 & 77.76$\pm$0.14 & 84.04$\pm$0.05 & 76.86$\pm$0.48 & 96.17$\pm$0.12 & 92.59$\pm$0.19 \\
\midrule
\textbf{TEAM} & \textbf{53.92$\pm$0.38} & \textbf{49.06$\pm$0.54} & \textbf{84.09$\pm$0.56} & \textbf{81.47$\pm$0.74} & \textbf{88.18$\pm$0.28} & \textbf{80.26$\pm$0.16} & \textbf{97.52$\pm$0.06} & \textbf{93.56$\pm$0.69} \\
\bottomrule
\end{tabular}}
\end{minipage}

%% file: sections/table_lp.tex
\begin{minipage}{\textwidth}
\centering
\captionof{table}{Link prediction on four MAG benchmarks (\%). Best results are
bold and second-best results are underlined.}
\label{tab:lp_main}
\fontsize{7.2}{7.8}\selectfont
\setlength{\tabcolsep}{2.8pt}
\renewcommand{\arraystretch}{0.70}
\resizebox{0.94\textwidth}{!}{%
\begin{tabular}{l cc cc cc cc}
\toprule
\multirow{2}{*}{Method}
& \multicolumn{2}{c}{TN}
& \multicolumn{2}{c}{Bili-Movie}
& \multicolumn{2}{c}{KU}
& \multicolumn{2}{c}{QB} \\
\cmidrule(lr){2-3}\cmidrule(lr){4-5}\cmidrule(lr){6-7}\cmidrule(lr){8-9}
& MRR & Hits@3 & MRR & Hits@3 & MRR & Hits@3 & MRR & Hits@3 \\
\midrule
MLP & 33.95$\pm$0.25 & 41.39$\pm$0.24 & 37.82$\pm$0.38 & 48.91$\pm$0.67 & 31.54$\pm$0.53 & 38.41$\pm$0.75 & 39.28$\pm$0.58 & 49.21$\pm$0.60 \\
\midrule
GCN & 47.60$\pm$0.39 & 65.71$\pm$0.34 & 47.18$\pm$0.71 & 64.99$\pm$0.47 & 55.80$\pm$0.38 & 74.54$\pm$0.34 & 59.10$\pm$0.66 & 77.28$\pm$0.73 \\
GAT & 47.09$\pm$0.66 & 64.21$\pm$0.80 & 47.59$\pm$0.49 & 64.59$\pm$0.79 & \underline{56.10$\pm$0.46} & \underline{75.50$\pm$0.59} & 59.07$\pm$0.37 & 77.84$\pm$0.53 \\
GraphSAGE & \underline{50.10$\pm$0.29} & \underline{67.10$\pm$0.09} & 47.60$\pm$0.27 & 63.05$\pm$0.46 & 55.86$\pm$0.14 & 70.51$\pm$0.48 & 61.47$\pm$0.51 & 76.78$\pm$0.54 \\
\midrule
Graphormer & 50.05$\pm$0.53 & 66.67$\pm$0.28 & 47.70$\pm$0.34 & 64.13$\pm$0.57 & 51.93$\pm$0.55 & 66.01$\pm$0.21 & \underline{62.13$\pm$0.50} & \underline{78.82$\pm$0.33} \\
DUALFormer & 44.78$\pm$0.36 & 60.57$\pm$0.93 & 45.05$\pm$0.28 & 62.37$\pm$0.69 & 53.14$\pm$0.18 & 71.16$\pm$0.30 & 57.24$\pm$0.16 & 74.42$\pm$0.40 \\
\midrule
MMBT & 47.67$\pm$0.51 & 63.65$\pm$0.37 & 47.23$\pm$0.38 & 62.86$\pm$0.42 & 43.70$\pm$1.41 & 54.70$\pm$2.40 & 47.55$\pm$2.47 & 59.64$\pm$2.32 \\
VLMo & 48.77$\pm$0.88 & 64.06$\pm$0.68 & 45.89$\pm$0.14 & 61.00$\pm$0.48 & 50.91$\pm$0.58 & 65.89$\pm$0.22 & 58.02$\pm$1.29 & 75.99$\pm$1.08 \\
\midrule
MMGCN & 44.78$\pm$0.67 & 62.74$\pm$1.65 & \underline{48.04$\pm$0.14} & 63.41$\pm$0.31 & 53.87$\pm$1.02 & 69.65$\pm$0.85 & 56.81$\pm$0.56 & 75.59$\pm$0.09 \\
DGF & 41.70$\pm$0.47 & 55.41$\pm$0.81 & 33.25$\pm$0.90 & 43.59$\pm$1.15 & 33.61$\pm$1.06 & 42.05$\pm$1.54 & 53.60$\pm$0.69 & 70.10$\pm$0.68 \\
DMGC & 44.65$\pm$0.19 & 61.00$\pm$0.32 & 44.35$\pm$0.50 & 59.17$\pm$0.57 & 44.73$\pm$0.10 & 56.97$\pm$0.51 & 57.62$\pm$0.90 & 74.92$\pm$1.14 \\
LGMRec & 50.09$\pm$0.67 & 67.03$\pm$0.64 & 47.81$\pm$0.17 & \underline{65.29$\pm$0.61} & 55.27$\pm$0.25 & 71.84$\pm$0.48 & 60.69$\pm$0.63 & 78.31$\pm$0.45 \\
\midrule
\textbf{TEAM} & \textbf{50.54$\pm$0.39} & \textbf{67.22$\pm$0.13} & \textbf{50.09$\pm$0.45} & \textbf{65.63$\pm$0.09} & \textbf{58.28$\pm$0.06} & \textbf{75.70$\pm$0.14} & \textbf{62.43$\pm$0.19} & \textbf{80.44$\pm$0.30} \\
\bottomrule
\end{tabular}}
\end{minipage}

%% file: sections/table_multimodal.tex
\begin{minipage}[t]{\textwidth}
\vspace{0pt}
\centering
\captionof{table}{Results on multimodal tasks. Best results are bold and
second-best results are underlined.}
\label{tab:generation_main}
\begin{minipage}{0.88\textwidth}
\centering
\fontsize{8.3}{9.2}\selectfont
\renewcommand{\arraystretch}{0.92}
\begin{tabular*}{\linewidth}{@{\extracolsep{\fill}}l ccc cc@{}}
\toprule
\multirow{2}{*}{Method}
& \multicolumn{3}{c}{G2Text: Flickr30k}
& \multicolumn{2}{c}{G2Image: SemArt} \\
\cmidrule(lr){2-4}\cmidrule(lr){5-6}
& BLEU-4 & ROUGE-L & CIDEr & CLIP & DINOv2 \\
\midrule
MLP & 5.90$\pm$0.18 & 29.16$\pm$0.31 & 36.75$\pm$0.74 & 75.82$\pm$0.11 & 61.49$\pm$0.41 \\
\midrule
GCN & \underline{6.11$\pm$0.15} & 29.45$\pm$0.27 & \underline{39.23$\pm$0.86} & 75.73$\pm$0.43 & 61.23$\pm$0.52 \\
GAT & 5.86$\pm$0.39 & 28.50$\pm$1.01 & 35.68$\pm$4.39 & \underline{76.09$\pm$0.18} & 61.43$\pm$0.04 \\
GraphSAGE & 5.83$\pm$0.17 & 28.52$\pm$0.29 & 36.27$\pm$0.91 & 75.73$\pm$0.12 & 60.94$\pm$0.65 \\
\midrule
Graphormer & 5.91$\pm$0.13 & 29.26$\pm$0.33 & 38.42$\pm$1.05 & 75.86$\pm$0.51 & 61.65$\pm$0.51 \\
DUALFormer & 5.25$\pm$0.55 & 28.06$\pm$0.99 & 32.06$\pm$2.72 & 76.02$\pm$0.08 & 61.53$\pm$0.17 \\
\midrule
MMBT & 5.56$\pm$0.20 & 28.43$\pm$0.41 & 33.68$\pm$0.97 & 75.96$\pm$0.12 & 61.15$\pm$0.37 \\
VLMo & 6.06$\pm$0.16 & \underline{29.57$\pm$0.34} & 38.61$\pm$0.88 & 75.66$\pm$0.22 & 61.29$\pm$0.15 \\
\midrule
MMGCN & 5.99$\pm$0.34 & 28.97$\pm$0.07 & 36.68$\pm$1.43 & 76.02$\pm$0.24 & 61.25$\pm$0.89 \\
DGF & 5.88$\pm$0.12 & 28.77$\pm$0.07 & 36.45$\pm$1.28 & 75.84$\pm$0.11 & \underline{61.68$\pm$0.13} \\
DMGC & 5.69$\pm$0.22 & 27.84$\pm$0.44 & 30.43$\pm$1.21 & 75.70$\pm$0.14 & 61.12$\pm$0.56 \\
LGMRec & 5.85$\pm$0.15 & 28.79$\pm$0.75 & 36.37$\pm$2.36 & 75.87$\pm$0.23 & 61.45$\pm$0.70 \\
\midrule
\textbf{TEAM} & \textbf{6.24$\pm$0.28} & \textbf{29.68$\pm$0.22} & \textbf{40.89$\pm$1.76} & \textbf{76.13$\pm$0.11} & \textbf{61.73$\pm$0.46} \\
\bottomrule
\end{tabular*}
\end{minipage}
\end{minipage}

%% file: sections/tables_efficiency.tex
\begin{minipage}{\textwidth}
\centering
\scriptsize
\setlength{\tabcolsep}{3pt}
\renewcommand{\arraystretch}{0.55}
\captionof{table}{Full-graph time and peak working-space upper bounds.}
\label{tab:complexity}
\resizebox{0.98\textwidth}{!}{%
\begin{tabular}{c c c c}
\toprule
Method & Training complexity & Inference complexity & Space complexity \\
\midrule
MLP & $\mathcal{O}(|V|d^2)$ & $\mathcal{O}(|V|d^2)$ & $\mathcal{O}(|V|d)$ \\
GCN & $\mathcal{O}(L|E|d+L|V|d^2)$ & $\mathcal{O}(L|E|d+L|V|d^2)$ & $\mathcal{O}(L|V|d+|E|)$ \\
GAT & $\mathcal{O}(L(|V|d^2+|E|d))$ & $\mathcal{O}(L(|V|d^2+|E|d))$ & $\mathcal{O}(L|V|d+|E|)$ \\
GraphSAGE & $\mathcal{O}(L|E|d+L|V|d^2)$ & $\mathcal{O}(L|E|d+L|V|d^2)$ & $\mathcal{O}(L|V|d+|E|)$ \\
MMGCN & $\mathcal{O}(ML|E|d)$ & $\mathcal{O}(ML|E|d)$ & $\mathcal{O}(ML|V|d)$ \\
DGF & $\mathcal{O}(|E|d+|V|d^2)$ & $\mathcal{O}(|E|d)$ & $\mathcal{O}(|V|d+|E|)$ \\
DMGC & $\mathcal{O}(|V|^2d+K|V|d)$ & $\mathcal{O}(|V|^2d)$ & $\mathcal{O}(|V|^2+Kd)$ \\
LGMRec & $\mathcal{O}(L|E|d+M|V|d)$ & $\mathcal{O}(L|E|d+M|V|d)$ & $\mathcal{O}((M+L)|V|d+|E|)$ \\
Graphormer & $\mathcal{O}(L|V|^2d+L|V|d^2)$ & $\mathcal{O}(L|V|^2d+L|V|d^2)$ & $\mathcal{O}(|V|^2+L|V|d)$ \\
DUALFormer & $\mathcal{O}(|V|+|E|)$ & $\mathcal{O}(|V|+|E|)$ & $\mathcal{O}(|V|d+|E|+d^2)$ \\
VLMo & $\mathcal{O}(|V|(LS^2d+LSd^2))$ & $\mathcal{O}(|V|(LS^2d+LSd^2))$ & $\mathcal{O}(|V|S^2+L|V|Sd)$ \\
MMBT & $\mathcal{O}(|V|(LS^2d+LSd^2))$ & $\mathcal{O}(|V|(LS^2d+LSd^2))$ & $\mathcal{O}(|V|S^2+L|V|Sd)$ \\
\midrule
\textbf{TEAM}
  & $\mathcal{O}\!\left(|V|C+|V|^2d_0+L|V|(S^2d+Sd^2)\right)$
  & $\mathcal{O}\!\left(|V|C+|V|^2d_0+L|V|(S^2d+Sd^2)\right)$
  & $\mathcal{O}\!\left(P_\theta+|E|+|V|(d_0+C+LSd)+|V|^2+L|V|hS^2\right)$ \\
\bottomrule
\end{tabular}}
\end{minipage}

%% file: sections/table_efficiency_measured.tex
\begin{minipage}{\columnwidth}
\centering
\fontsize{6.5}{7.0}\selectfont
\setlength{\tabcolsep}{2.5pt}
\renewcommand{\arraystretch}{0.45}
\captionof{table}{Per-epoch time and peak memory on Movies.}
\label{tab:efficiency}
\resizebox{\linewidth}{!}{%
\begin{tabular}{c c c c}
\toprule
Family & Method & Epoch time (s) & Memory (MiB) \\
\midrule
Feature-only & MLP & 1.62 & 516 \\
GNN & GCN & 1.67 & 500 \\
GNN & GAT & 2.55 & 4770 \\
GNN & GraphSAGE & 2.18 & 3824 \\
GNN & MMGCN & 1.95 & 514 \\
GNN & DGF & 6.59 & 46910 \\
GNN & DMGC & 2.54 & 3132 \\
GNN & LGMRec & 2.18 & 836 \\
Transformer & Graphormer & 20.55 & 48052 \\
Transformer & DUALFormer & 1.67 & 776 \\
Transformer & VLMo & 14.09 & 2890 \\
Transformer & MMBT & 2.44 & 2122 \\
\midrule
Transformer & \textbf{TEAM} & 7.53 & 1626 \\
\bottomrule
\end{tabular}}
\end{minipage}

%% file: sections/ethical_considerations.tex
\noindent\begin{minipage}{\linewidth}
\setcounter{section}{5}
\section{Ethical Considerations}

TEAM uses existing benchmarks and collects no new human-subject data.
Multimodal attributes may encode sensitive information, while semantic
supplementation may amplify frequent entities, hubs, or spurious correlations.
Deployments should respect licenses and access controls, audit relevant groups
and classes, and prevent disclosure of retrieved content.  Candidate pools
must follow the evaluated graph's split and access-control protocol so context
construction cannot access held-out or unauthorized entities.  Our benchmark
results do not establish fairness, privacy, or safety guarantees for high-stakes
deployment.
\end{minipage}